\documentclass[a4paper, 10pt, conference]{ieeeconf}      
\usepackage{FG2025}
\usepackage{amsmath} 
\usepackage{multirow}
\usepackage{graphicx}
\usepackage{booktabs} 
\usepackage[ruled,vlined]{algorithm2e}
\PassOptionsToPackage{hyphens}{url}\usepackage{hyperref}
\FGfinalcopy 
\usepackage{fancyhdr}

\IEEEoverridecommandlockouts                              
\def\FGPaperID{326} 

\title{\LARGE \bf
DiffProb: Data Pruning for Face Recognition
}

\author{\parbox{16cm}{\centering
    {\large Eduarda Caldeira$^1$, Jan Niklas Kolf$^{1,2}$, Naser Damer$^{1,2}$ and Fadi Boutros$^1$ }\\
    {\normalsize
    $^{1}$Fraunhofer IGD, Germany, $^{2}$Department of Computer Science, TU Darmstadt, Germany\\
   }}
    \thanks{This research work has been funded by the German Federal Ministry of Education and Research and the Hessian Ministry of Higher Education, Research, Science and the Arts within their joint support of the National Research Center for Applied Cybersecurity ATHENE.}
}

\begin{document}

\ifFGfinal
\thispagestyle{empty}
\pagestyle{empty}
\else
\author{Anonymous FG2025 submission\\ Paper ID 326 \\}
\pagestyle{plain}
\fi
\maketitle

\thispagestyle{fancy}
\renewcommand{\headrulewidth}{0pt}
\fancyhf{}

\begin{abstract}
 Face recognition models have made substantial progress due to advances in deep learning and the availability of large-scale datasets. However, reliance on massive annotated datasets introduces challenges related to training computational cost and data storage, as well as potential privacy concerns regarding managing large face datasets. This paper presents DiffProb, the first data pruning approach for the application of face recognition.  
 DiffProb assesses the prediction probabilities of training samples within each identity and prunes the ones with identical or close prediction probability values, as they are likely reinforcing the same decision boundaries, and thus contribute minimally with new information.
 We further enhance this process with an auxiliary cleaning mechanism to eliminate mislabeled and label-flipped samples, boosting data quality with minimal loss. Extensive experiments on CASIA-WebFace with different pruning ratios and multiple benchmarks, including LFW, CFP-FP, and IJB-C, demonstrate that DiffProb can prune up to 50\% of the dataset while maintaining or even, in some settings, improving the verification accuracies.
 Additionally, we demonstrate DiffProb’s robustness across different architectures and loss functions. Our method significantly reduces training cost and data volume, enabling efficient face recognition training and reducing the reliance on massive datasets and their demanding management. The code, pretrained models, and pruned datasets are publicly released: \url{https://github.com/EduardaCaldeira/DiffProb}.
 \end{abstract}

\section{Introduction}
In recent years, face recognition (FR) systems have achieved remarkable progress, primarily driven by advancements in deep neural architectures \cite{He2015DeepRL,DBLP:conf/iclr/DosovitskiyB0WZ21}, the availability of large-scale training datasets, and the development of margin-based SoftMax loss functions  \cite{DBLP:conf/cvpr/Kim0L22,DBLP:conf/cvpr/HuangWT0SLLH20}. These innovations have significantly boosted the performance of FR models, making them integral to applications ranging from surveillance to on-device identification. However, achieving state-of-the-art (SOTA) results typically requires training deep neural networks (DNNs) on massive datasets such as CASIA-WebFace (0.5M images)  \cite{DBLP:journals/corr/YiLLL14a}, VGGFace (2.6M images) \cite{wang2018cosfacelargemargincosine}, VGGFace2 (3.3M images) \cite{8373813}, WebFace260M (260M images) \cite{DBLP:conf/cvpr/ZhuHDY0CZYLD021}, and MS1M-CelebA (10M images) \cite{DBLP:conf/eccv/GuoZHHG16}. While such development is very effective in achieving SOTA FR performances \cite{DBLP:conf/cvpr/Kim0L22,DBLP:conf/cvpr/HuangWT0SLLH20}, the reliance on massive training data introduces several challenges, including increased training time, computational overhead, large data storage and sharing requirements, and growing concerns over data privacy and ethical implications \cite{SyntheticFRSurvay,DBLP:journals/cviu/GuoZ19}. These challenges, though not unique to FR, are particularly pressing in this domain due to the sensitive nature of biometric data \cite{lirias3838501}.

Similar problems appeared in general computer vision problems, such as object classification. In these cases, to reduce the reliance on massive training datasets, the computer vision research community has conducted several studies \cite{DBLP:conf/iclr/TonevaSCTBG19,DBLP:conf/nips/PaulGD21,DBLP:conf/iclr/0006XP0S023} that explored and proposed data pruning approaches, a process of identifying and retaining only the most informative or representative training samples. Effective pruning can significantly reduce dataset size while maintaining, or even improving in some cases, model performance \cite{DBLP:conf/cvpr/HeY0Z22}. 
This enables more efficient training and better scalability, especially in resource-constrained settings where on-device training might be needed. Early pruning approaches focused on core-set selection, aiming to retain a subset of samples that preserves the decision boundary of the original data. More recent techniques leverage influence functions \cite{DBLP:conf/iclr/0006XP0S023}, uncertainty estimation \cite{DBLP:conf/cvpr/HeY0Z22}, and training dynamics  \cite{DBLP:conf/cvpr/ZhangDLXZ24} to guide the pruning process. For instance, Paul et al. \cite{DBLP:conf/nips/PaulGD21} proposed identifying impactful examples early in training based on their contribution to generalization. Yang et al. \cite{DBLP:conf/iclr/0006XP0S023} employed influence estimation to prune data without compromising performance across architectures, while He et al. \cite{DBLP:conf/cvpr/HeY0Z22} introduced Dynamic Uncertainty (DynUnc), which discards low-contribution samples by tracking prediction uncertainty over time. These architecture-agnostic strategies highlight the potential of data pruning approaches to efficient training.
Despite these advances, the application of data pruning specifically to face recognition remains underexplored. To the best of our knowledge, none of the prior works \cite{DBLP:conf/nips/PaulGD21,DBLP:conf/cvpr/HeY0Z22}  have explicitly investigated data pruning strategies for face recognition datasets.

This work aims to address the issue of large data requirements, along with its management cost, and the lack of data pruning consideration for FR by introducing a novel data pruning approach, DiffProb. DiffProb prunes face recognition training datasets, enabling training FR models with smaller annotated face data.
DiffProb prunes samples with identical or very close prediction probabilities, theorizing the removal samples with redundant or minimum novel contributions toward the FR model's learning process. 
Intuitively, samples with highly similar prediction probabilities are likely to reinforce the same decision boundaries, and thus contribute minimal new information. 
We empirically proved that DiffProb is highly effective across several training settings, consistently pruning the FR datasets to a significant degree (up to 50\%) while keeping or even increasing the performance levels achieved with the complete dataset. 
The reduction in training dataset size also translates into a lower training cost, both in terms of computation and time requirements.
We further address the issue of mislabeled and label-flipped samples in FR datasets by proposing an auxiliary dataset cleaning mechanism that removes samples whose predicted class does not correspond to their ground truth label. By removing only 1.13\% of the CASIA-WebFace dataset, this mechanism was able to significantly increase the FR verification accuracies on the challenging benchmark IARPA Janus Benchmark–C (IJB-C) by 14.07 percentage points.


\begin{table}[t!]
\scriptsize
    \centering
        \caption{Training datasets commonly used to train FRs. The recent trend is to utilize massive datasets with millions of samples.}
    \begin{tabular}{c|c|c|c|c}
        Dataset & \# Identities & \# Images & Images/ID & Publication \\ \hline \hline
        CelebFaces \cite{DBLP:conf/iccv/SunWT13} & 10k & 0.2M & 20 & ICCV 2013 \\
        CASIA-WebFace \cite{DBLP:journals/corr/YiLLL14a} & 10k & 0.5M & 47 & Arxiv 2014 \\
        DeepFace \cite{DBLP:conf/cvpr/TaigmanYRW14} &  4k & 4.4M & 1,100 & CVPR 2014 \\
Facebook \cite{DBLP:conf/cvpr/TaigmanYRW15} &  10M & 500M & 50 & CVPR 2015 \\
FaceNet \cite{DBLP:conf/cvpr/SchroffKP15} &  8M & 200M & 25 & CVPR 2015 \\
VGGFace \cite{wang2018cosfacelargemargincosine} & 2k & 2.6M & 1,000 & BMVC 2015 \\
MS1M \cite{DBLP:conf/eccv/GuoZHHG16} & 0.1M & 10M & 100 & ECCV 2016 \\
MS1M-IBUG \cite{DBLP:conf/cvpr/DengZZ17} & 85k & 3.8M & 45 & CVPRW 2017 \\
UMDFaces \cite{DBLP:conf/icb/BansalNCRC17} & 8k & 0.3M & 45 & IJCB 2017 \\
MegaFace2 \cite{DBLP:conf/cvpr/NechK17} & 0.6M & 4.7M & 7 & CVPR 2017 \\
VGGFace2 \cite{8373813} & 9k & 3.3M & 363 & FG 2018 \\
IMDB-Face \cite{DBLP:conf/eccv/WangCLHCQL18} & 59k & 1.7M & 29 & ECCV 2018 \\
MS1M-Glint \cite{deepglint_trillionpairs} & 87k & 3.9M & 44 & - \\
MS1MV2 \cite{DBLP:conf/cvpr/DengGXZ19} & 85k & 5.8M & 68 & CVPR 2019 \\
MillionCelebs \cite{DBLP:conf/cvpr/ZhangDWHLZW20} &  0.6M & 18.8M & 30 &  CVPR 2020 \\
WebFace260M \cite{DBLP:conf/cvpr/ZhuHDY0CZYLD021} & 4M & 260M & 65 & CVPR 2021 \\
WebFace42M \cite{DBLP:conf/cvpr/ZhuHDY0CZYLD021} & 2M & 42M & 21 & CVPR 2021
    \end{tabular}
    \label{tab:databases}
    \vspace{-3mm}
\end{table}

\section{Related Work}
\label{sec:related_work}
\subsection{Face Recognition Training Datasets}
Recent SOTA FR models \cite{DBLP:conf/eccv/BoutrosSD24, wang2018cosfacelargemargincosine, DBLP:journals/ivc/ChettaouiDB25, DBLP:conf/cvpr/BoutrosDKK22, DBLP:conf/cvpr/DengGXZ19} proposed to train very deep network architectures on massive training datasets, such as CASIA-WebFace (0.5M images) \cite{DBLP:journals/corr/YiLLL14a}, VGGFace2 (3.3M images) \cite{8373813}, MS1MV2 (5.8M images) \cite{DBLP:conf/cvpr/DengGXZ19} and WebFace42M (42M images) \cite{DBLP:conf/cvpr/ZhuHDY0CZYLD021}. A detailed list of training datasets commonly used in the literature to train FR is provided in Table \ref{tab:databases}, highlighting the previously described trend. While this practice has enabled the progress of FR research at a fast pace, factors such as model training time, computational overhead, and data sharing and storage challenges could significantly benefit from reducing the size of FR training datasets. 
This is a common problem in different applications of machine learning. In such applications, previous works proposed to mitigate these challenges through data pruning solutions \cite{DBLP:conf/iclr/TonevaSCTBG19, DBLP:conf/nips/PaulGD21, DBLP:conf/cvpr/HeY0Z22, DBLP:conf/cvpr/ZhangDLXZ24, DBLP:conf/iclr/0006XP0S023}, which aim to reduce the number of training samples, and thus the aforementioned challenges. 
These previous data pruning approaches \cite{DBLP:conf/iclr/TonevaSCTBG19, DBLP:conf/nips/PaulGD21, DBLP:conf/cvpr/HeY0Z22, DBLP:conf/cvpr/ZhangDLXZ24, DBLP:conf/iclr/0006XP0S023} are commonly applied to general machine learning tasks. However, none of these works evaluate or propose a solution to prune face datasets, motivating our DiffProb to enable FR training with few annotated data.


\subsection{Data Pruning}
In this section, we highlight the recent works that proposed data pruning for general computer vision tasks. 
Toneva~\textit{et al.}~\cite{DBLP:conf/iclr/TonevaSCTBG19} defined a forgetting score for each sample by determining the number of times a sample transitions from correctly classified to misclassified during training. Their results revealed that some samples were forgotten on multiple occasions while others rarely transitioned to being misclassified. They observed that this last group of samples did not have a significant impact on the training accuracy, allowing for efficient dataset pruning. 
Paul~\textit{et al.}~\cite{DBLP:conf/nips/PaulGD21} proposed a pruning strategy based on an importance score.
The importance score of a sample is determined based on the expected impact of removing it on the loss gradient norm over the first training epochs, allowing for pruning the dataset without requiring to fully train the network beforehand. 
He~\textit{et al.}~\cite{DBLP:conf/cvpr/HeY0Z22} presents an uncertainty-based pruning method that separates the samples according to the level of constancy presented by the model in their classification. Their method is based on the fact that pruning methods should aim at removing both the extremely easy and extremely hard samples, due to their redundancy and high complexity level, respectively. When analyzed over the training process, these samples are expected to maintain consistent predictions because they are either learned in the early stages of training or not learned at all. Hence, their proposed DynUnc approach prunes the samples with low prediction uncertainty during the training process. To that end, the prediction standard deviation is determined for a sliding window through training epochs and the obtained values are averaged across all the considered sliding windows, resulting in the final uncertainty scores. Similar to DynUnc \cite{DBLP:conf/cvpr/HeY0Z22}, Zhang~\textit{et al.}~\cite{DBLP:conf/cvpr/ZhangDLXZ24} proposed to maximize the variance of the kept samples using information about the gradient magnitude. In particular, they project each sample's gradient in a gradient vector representing the whole dataset.  This makes the effect of each sample's contribution on the overall dataset panorama more evident. Furthermore, the samples that contribute more to the original training process are also given more importance when retraining the model with the pruned dataset, which better reflects the importance of each sample from the original training dataset.  
Yang~\textit{et al.}~\cite{DBLP:conf/iclr/0006XP0S023} optimized the concept of influence functions \cite{hampel1974influence} to account for the joint effect of pruning multiple samples. This constitutes an interesting approach from the data pruning problem's perspective since samples that seem to contain critical information when analyzed individually might be deemed redundant when their combined effect is considered. They demonstrated that it is possible to use the dataset pruned based on the metrics extracted from a model to train a distinct architecture without significant performance drops \cite{DBLP:conf/iclr/0006XP0S023}. 
The generalizability of pruning methods, as demonstrated in \cite{DBLP:conf/iclr/0006XP0S023}, is of the utmost importance as it reveals the absence of a need to re-prune the dataset when using fewer data in different training settings, as later shown in Sections \ref{sec:cross_loss} and \ref{sec:cross_net}. 

While previous works considered only pruning datasets designed for general computer vision tasks, such as ImageNet \cite{DBLP:conf/cvpr/DengDSLL009} and CIFAR \cite{krizhevsky2009learning}, which typically require a much higher level of feature granularity as discussed in \cite{DBLP:conf/eccv/ZeilerF14}, this is the first work to propose an approach to prune face datasets.


\section{Methodology}
\label{sec:methods}
\vspace{-1mm}
This section describes our data pruning framework, DiffProb, which aims to prune redundant samples within each identity in the training dataset. Our approach is based on the hypothesis that samples with identical or very similar prediction probabilities (according to a tunable threshold $t$) are pushed with a comparable degree toward their respective class center during training. As a result, such samples are considered redundant, and some of them can be removed without significantly affecting the model's learning dynamics. To ensure that no class is underrepresented, a minimum number of samples is enforced per identity. Complementing this, an auxiliary cleaning mechanism is proposed to preemptively eliminate mislabeled and label-flipped instances, identified as samples misclassified by the FR model. This two-stage pipeline of cleaning followed by pruning results in significantly smaller yet highly informative datasets.


\subsection{DiffProb Pruning}
\label{sec:local_pruning}



Let $D$ be a dataset containing $N$ samples of $C$ distinct classes. Let $x_i$ be the $i^{th}$ sample ($i\in [1, N]$) and $y_i$ be its corresponding ground truth label ($1\leq y_i\leq C$). The subset of samples belonging to identity $id\in [1, C]$ can be defined as:

\begin{equation}
    D_{id}=\{x_i \in D \mid\ y_i = id\},
\end{equation}
and the number of samples belonging to identity $id$ is defined as $n_{id} = \left| D_{id} \right|, \, n_{id} \leq N$. 

The proposed DiffProb mechanism is designed to prune samples with very similar prediction probabilities within each identity. These predictions are obtained by passing all input samples $x_i$ through a pre-trained FR model, $f$. The resultant probability vector is obtained by applying the softmax activation on the output of the classification layer of $f$:

\begin{equation}
    v_{yi} = \frac{e^{z_{yi}}}{\sum_{j=1}^{C}e^{z_j}},
\end{equation}

where $v_{yi}$ is the $yi^{th}$ entry of the final probability vector and $z_{j}=f(x_i)_{j}$ is the logit output by passing $x_i$ through $f$ and corresponds to the $yi^{th}$ entry of the vector outputted by $f$ when fed sample $x_i$. The different $v_j$ values in this vector correspond to the probabilities determined for each class label, including the one ($v_{y_i}$) corresponding to $y_i$. 
Given $f$ trained with margin penalty Softmax loss, in this paper, CosFace, $z_{yi}=cos(\theta_{y_i})-m$ and $z_j=cos(\theta_{j})$.  $\theta_{y_i}$ is the angle between the features $f(x_{i})$ and the $y_i$-th class center $w_{y_i}$. {f($x_{i}) \in R^d$}  is the deep feature embedding of the last fully connected layer of size $d$. $w_{y_i}$ is the $y_i$-th column of weights $W \in R^d_C$ of the classification layer.
$\theta_{y_i}$ is defined as ${f(x_i)w^T_{y_i}}=\Vert f(x_i) \Vert \Vert w_{y_i} \Vert cos(\theta_{y_i})$ \cite{DBLP:conf/cvpr/LiuWYLRS17}. The weights and the feature norms are fixed to $\Vert w_{y_i} \Vert=1$ and $ \Vert f(x_i) \Vert =1$, respectively, using $l_2$ normalization as defined in \cite{DBLP:conf/cvpr/LiuWYLRS17,wang2018cosfacelargemargincosine}. The decision boundary, in this case, depends on the cosine between $f(x_{i})$ and $w_{y_i}$ and the predicted probabilities $v_{yi}$ represent similarity of $f(x_i)$ being close to its class center $w_{yi}$ with respect to other class centers $w_j$.   
During the training, 
$m>0$ is an additive cosine margin proposed by CosFace \cite{wang2018cosfacelargemargincosine} to enhance the intra-class compactness and inter-class discrepancy. During the inference phase, to extract prediction probabilities of pretrained $f$, $m$ is set to $0$.
We note this probability as $p(x_i)$ and utilize it as the pruning criterion for our DiffProb approach ($p(x_i)=v_{y_i}$). 
We then utilize a threshold $t$ to determine if two samples $x_i$ and $x_j$ of the same class label with probability predictions $p(x_i)$ and $p(x_j)$, respectively, are similar to each other and, thus, they are likely reinforcing the same decision boundaries, and thus contribute minimal
new information.
Since pruning is performed within each identity, each value of $t$ results in pruned identity subsets $D_{id}(t)$ whose union corresponds to the final pruned dataset, $D(t)$:

\begin{equation}
    D(t) = \bigcup_{id=1}^{C} D_{id}(t).
\end{equation}

To ensure that no class is underrepresented, a minimum number of samples $n_{min}$ is also enforced per identity\footnote{Note that there are identities with $n_{id}\leq n_{min}$. The samples belonging to these identities do not undergo the pruning process, as described in Algorithm \ref{alg:local}.}.

Our complete DiffProb pruning process is described in Algorithm \ref{alg:local}. We start by sorting the samples of each identity $id$ in ascending order based on $p(x_i),\; x_i\in D_{id}$:

\begin{align}
    O_{id} & = (x_{(1)}, x_{(2)}, \dots, x_{(n_{id})}),\; \\ 
    & x_{(i)} \in D_{id} \notag 
            \land p(x_{(i)}) \leq p(x_{(j)}), \; \forall i < j.
\end{align}

The proposed pruning criterion builds $D_{id}(t)$ gradually by scanning $O_{id}$ sample by sample and individually deciding whether to include  $x_{(i)} \in O_{id}$ in the pruned subsets. For each sample $x_{(i)}$, our pruning criterion considers it redundant if its effect $p(x_{(i)})$ is deemed similar to the effect of a sample already included in the identity's pruned dataset\footnote{Note that this requires $D_{id}(t)$ to include at least one sample when the first comparison is performed. Hence, $D_{id}(t)$ is initialized as $\{x_{(n_{id})}\}$, as described in Algorithm \ref{alg:local}.}. Since the samples in $O_{id}$ are ordered by their $p(.)$ values, this criterion only requires the comparison with the last sample added to $D_{id}(t)$, $x_{(l)}$. The concept of similarity is determined by the selected threshold $t$, where a larger $t$ allows for more aggressive pruning and a smaller $t$ results in a more lenient pruning process:

\begin{equation}
D_{id}(t) =
\begin{cases} 
    D_{id}(t) \cup x_{(k)}, & \text{if } p(x_{(l)}) - p(x_{(k)}) > t \\
    D_{id}(t), & \text{otherwise}
\end{cases}
\label{eq:prune_local}.
\end{equation}
 
It should be noted that after analyzing all the samples in $O_{id}$, the resultant $D_{id}(t)$ may contain fewer samples than $n_{min}$. In this scenario, the value of $t$ is slightly decreased by 1\% of the original $t$ value, allowing for a less constrained filtering process, and $O_{id}$ is pruned from scratch using the updated value. This process ensures that each class in the pruned dataset contains at least a minimum of $n_{min} >1$ samples. 
This process is repeated until $ \left| D_{id}(t) \right| \geq n_{min}$, as described in Algorithm \ref{alg:local}. 




\subsection{Cleaning Mechanism}
\label{sec:cleanining}
\begin{figure}[t!]
    \centering
    \includegraphics[width=0.75\linewidth]{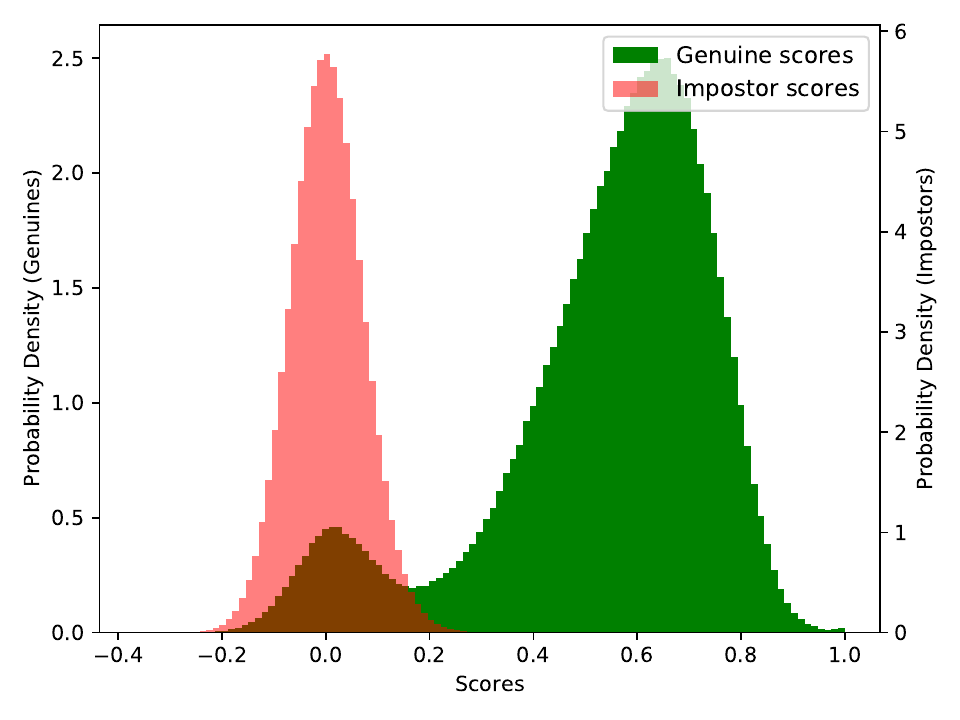}
\vspace{-3mm}    \caption{Histograms of the genuine and impostor score distributions of CASIA-WebFace \cite{DBLP:journals/corr/YiLLL14a} when evaluated by an FR model trained on MS1MV2 \cite{DBLP:conf/cvpr/DengGXZ19}. One can notice that the genuine score distribution (green) contains a left wing of comparison scores that highly overlapped with the impostor distribution, highlighting the presence of mislabeled and label-flipped samples in the CASIA-WebFace dataset.}
    \label{fig:casia_dist}
    \vspace{-2mm}
\end{figure}

\vspace{-1mm}
Apart from effectively identifying and removing redundant samples, an ideal pruning mechanism should consider mislabeled or label-flipped samples as more dominant samples to be pruned. 
Datasets such as CASIA-WebFace \cite{DBLP:journals/corr/YiLLL14a} often contain mislabeled and label-flipped samples \cite{DBLP:conf/iccv/WangWSWM19}, as can be easily verified by analyzing the left wing of the genuine scores distribution of CASIA-WebFace \cite{DBLP:journals/corr/YiLLL14a} when evaluated by an FR model trained on MS1MV2 \cite{DBLP:conf/cvpr/DengGXZ19} (Figure \ref{fig:casia_dist}). These samples contribute to degrading the performance of FR systems and should thus be eliminated by an efficient pruning method. 

To this end, we proposed to extend our DiffProb pruning method to identify first the mislabeled or label-flipped samples and then apply our pruning approach. Considering this, we propose an auxiliary data cleaning mechanism that aims to remove the mislabeled or label-flipped samples. Hence, our cleaning mechanism simply removes all the samples that are misclassified by the pretrained FR system. As later elaborated in Section \ref{sec:res_cleaning}, this results in the removal of 1.13\% of CASIA-WebFace's samples. Our DiffProb pruning method can then be more safely applied on top of the resultant clean dataset.

\begin{algorithm}
\small
\caption{DiffProb Data Pruning Algorithm}
    \KwIn{Original dataset $D$ with $C$ identities, threshold $t$, minimum number of samples per id $n_{min}$, prediction of the ground truth label probability for a sample by a pre-trained FR model $p(.)$}
    \KwOut{Pruned dataset $D(t)$}
    \For{$id=1$ \textbf{to} $C$}{
        Initialize $t_{dec} = 0$ \\
        Initialize $t_{step} = 0.01$ \\
        Initialize $D_{id} = \{ x_i \in D \mid y_i = id \}$ \\
        Initialize $N_{id} = \left| D_{id} \right|$ \\
        Initialize $O_{id} = (x_{(1)}, x_{(2)}, \dots, x_{(n_{id})}), \; \text{where}$ \\ $x_{(i)} \in D_{id}, \; p(x_{(1)}) \leq p(x_{(2)}) \leq \dots \leq p(x_{(n_{id})})$ \\
        \If{$n_{id} \leq n_{min}$}{$D_{id}(t)\gets D_{id}$
        }
        \Else{
            Initialize $D_{id}(t) = \{\}$ \\
            \While{$\left| D_{id}(t) \right| < n_{min}$}{
                Initialize $D_{id}(t) = \{ x_{(n_{id})} \}$ \\  
                Initialize $x_{(l)} = x_{(n_{id})}$ \\                
                \For{$x_{(k)} \in O_{id}, \; k = n_{id}-1$ \textbf{to} $1$}{
                    \If{$p(x_{(l)})-p(x_{(k)})>(1+t_{dec})\times t$}{$D_{id}(t) \gets D_{id}(t) \cup x_{(k)}$ \\ $x_{(l)} \gets x_{(k)}$
                    }
                }
                $t_{dec} \gets t_{dec} - t_{step}$
            }
        }
    }
    $D(t)=\bigcup_{id=1}^{C} D_{id}(t)$
\label{alg:local}
\end{algorithm}

\vspace{-1mm}
\section{Experimental Setup}
\label{sec:exp_setup}

\subsection{Baseline Pruning Methods}
Since our method is the first to address dataset pruning in the context of FR, there are no prior works that provide a direct comparison with our method. 
Thus, to evaluate our proposed DiffProb's effectiveness, we implemented two baseline methods widely used in general computer vision tasks \cite{DBLP:conf/cvpr/HeY0Z22}. The first implemented approach is random (Rand) pruning, where samples within each identity are randomly removed from the training dataset.  We further implement DynUnc \cite{DBLP:conf/cvpr/HeY0Z22} as our second baseline. This method was originally developed to perform dataset pruning for image classification tasks, and we adapted it for FR.
This method was selected due to its significant superiority when compared with other data pruning approaches \cite{DBLP:conf/cvpr/HeY0Z22}, making it the strongest baseline candidate among previously proposed methods.
DynUnc \cite{DBLP:conf/cvpr/HeY0Z22} tracks the prediction uncertainty of samples throughout the training process and identifies low-contribution samples by measuring the change in prediction probabilities over training iterations, under the assumption that samples with consistently low uncertainty provide limited learning value. Note that the original DynUnc \cite{DBLP:conf/cvpr/HeY0Z22} requires training a model from scratch to extract the predictions of each sample during the training process. Both Rand and DynUnc were conditioned to keep the same minimum number of samples per identity as our DiffProb strategy.
While there are no rules to select the minimum number of samples per class, we followed the common practice of 5-shot learning \cite{DBLP:conf/cvpr/GidarisK18} and set $n_{min}=5$.


\subsection{Face Recognition Training Setup}
The FR models presented in this paper use distinct architectures and loss functions. We start by training a model on the complete CASIA-WebFace dataset \cite{DBLP:journals/corr/YiLLL14a} to provide baseline results and obtain $p(x_i), \; x_i\in D$, required for our DiffProb (Section \ref{sec:methods}) and for the competitor (DynUnc \cite{DBLP:conf/cvpr/HeY0Z22}). This model uses ResNet-50 \cite{He2015DeepRL} as network architecture and is trained with the CosFace loss \cite{wang2018cosfacelargemargincosine}, with a margin penalty $m$ of 0.35 and the scale factor $s$ of 64, following \cite{wang2018cosfacelargemargincosine}. We utilized the model trained on the complete dataset to prune  CASIA-WebFace with different ratios ($\sim$ 25\%, 50\% and 75\%, which correspond to 
$\sim$ 360k, 240k and 120k samples, respectively) using our DiffProb and DynUnc. One should note that random pruning does not require access to a pretrained model. We then train 9 instances of ResNet50 from scratch using the exact training setup mentioned earlier in this section.
To evaluate the generalizability of our DiffProb over different loss functions and network architectures, we provided evaluation results for cross-loss and cross-network settings using the pruned datasets from previous experiments. 
Specifically, for the cross-loss experiments, the architecture was fixed as ResNet-50 \cite{He2015DeepRL}. The considered loss functions were the AdaFace loss \cite{DBLP:conf/cvpr/Kim0L22}, and the CurricularFace loss ($m=0.5$ and $s=64$) \cite{DBLP:conf/cvpr/HuangWT0SLLH20}. 
These loss functions were selected due to their competitive performances within SOTA FR \cite{DBLP:conf/cvpr/Kim0L22}.
For the cross-network scenario, the loss function is fixed as the CosFace loss ($m=0.35$ and $s=64$) \cite{wang2018cosfacelargemargincosine} and the considered architecture was ResNet-34 \cite{He2015DeepRL}. The remaining setup is the same for all performed experiments: the mini-batch size is set to 256 and the models are trained with a Stochastic Gradient Descent (SGD) optimizer with a momentum of 0.9 and a weight decay of 5e-4, following \cite{DBLP:conf/cvpr/DengGXZ19}.
The initial learning rate is set to 0.1 and is reduced by a factor of 10 after 8, 14, 20 and 25 epochs. 

\begin{figure*}[ht!]
    \centering
    \includegraphics[width=0.85\linewidth]{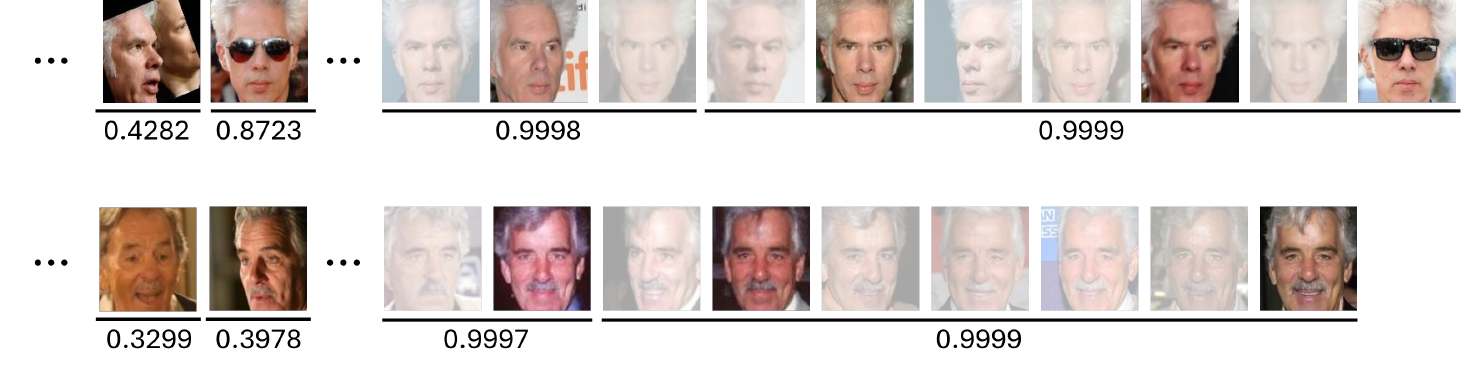}
    \vspace{-3mm}
    \caption{Visual representation of two distinct identities of the CASIA-WebFace dataset and their processing by our DiffProb pruning method. Translucent images correspond to the samples pruned when $t=0.00003$, which leads to pruning 25\% of the dataset samples across all identities. The values below each sample $x_i$ represent their $p(x_i)$ value, truncated to four decimal places. Note that samples assigned to the same truncated value correspond to different $p(x_i)$ values when full precision is considered and that the samples are ordered in ascending order of their full precision $p(x_i)$ from left to right. This justifies why some samples within the same range are pruned and others are not. It can also be observed that hard samples, such as the ones present in the left part of the figure, are not the first to be pruned by our DiffProb method, which is beneficial as they can contribute to the FR model learning process.
    }
    \label{fig:pruning}
\end{figure*}

\subsection{Evaluation Benchmarks and Metrics}
\label{sec:eval_benchmarks}
In this work, the FR verification performances are evaluated and reported as the verification accuracy on five benchmarks: Labeled Faces in the Wild LFW \cite{LFWDatabase}, AgeDb-30 \cite{AgeDB30Database}, Cross-Age LFW (CA-LFW) \cite{CALFWDatabase}, Celebrities in Frontal-Profile in the Wild (CFP-FP) \cite{CFPFPDatabase}, and Cross-Pose LFW (CP-LFW) \cite{CPLFWDatabase}, following their official evaluation protocol. In addition, we evaluated on the large-scale evaluation benchmark IARPA Janus Benchmark–C (IJB-C) \cite{DBLP:conf/icb/MazeADKMO0NACG18}. For IJB-C, we used the official 1:1 mixed verification protocol and reported the verification performance as True Acceptance Rates (TAR) at False Acceptance Rates (FAR) of 1e-4 and 1e-5 \cite{DBLP:conf/icb/MazeADKMO0NACG18}. 

\vspace{-1mm}
\section{Results and Discussion}
\label{sec:results}

\begin{table*}[ht!]
    \centering
        \caption{Performance of FR models trained on datasets pruned with different methods to three pruning percentages. All the considered models use the same architecture and loss function as the FR model used to prune the original dataset (ResNet-50 and CosFace).
    The best result for each benchmark and pruning percentage is marked in bold, showing that our DiffProb achieves the highest performance in most of the analyzed scenarios.}
    \begin{tabular}{cc||cccccc|cc}
     \multirow{2}{*}{\textbf{Method}} & \multirow{2}{*}{ \textbf{Kept Samples (\%)} } & \multirow{2}{*}{\textbf{LFW}} & \multirow{2}{*}{\textbf{CFP-FP}} & \multirow{2}{*}{\textbf{AgeDB}} & \multirow{2}{*}{\textbf{CA-LFW}} & \multirow{2}{*}{\textbf{CP-LFW}} & \multirow{2}{*}{\textbf{Avg}} & \multicolumn{2}{c}{\textbf{IJB-C}} \\
    & & & & & & & & \textbf{$10^{-5}$} & \textbf{$10^{-4}$} \\ \hline 
      - & 100 & 99.35	& 95.24 & 94.65 & 93.83 & 90.17 & 94.65 & 62.95	& 85.26 \\ \hline \hline
      \multirow{3}{*}{DynUnc \cite{DBLP:conf/cvpr/HeY0Z22}} & 75 & 99.30 & 94.90 & 94.27 & 93.75 & 89.02 & 94.25 & 72.66 & 86.01 \\
      & 50 & 99.18 & 92.43 & 93.35 & 93.20 & 86.77 & 92.99 & 67.90 & 82.56 \\
       & 25 & \textbf{98.77} & 86.97 & \textbf{90.15} & \textbf{91.43} & \textbf{82.57} & \textbf{89.98} &	\textbf{66.56} & \textbf{78.16} \\ 
      \hline
     \multirow{3}{*}{Rand} & 75 & 99.40 & 94.54 & 94.22 & 93.27 & 89.33	& 94.15 & 75.74 & 86.22\\
      & 50 & 99.30 & 93.30 & 93.40 & 92.83 & 87.97 & 93.36 & 75.69 & 84.59\\
       & 25 & 98.38 & 87.50 & 89.22 & 90.87 & 81.53 & 89.50 & 62.13 & 74.77\\ 
     \hline
      \multirow{3}{*}{DiffProb (Ours)} & 75 & \textbf{99.50}	& \textbf{95.47}	& \textbf{94.82}	& \textbf{93.87}	& \textbf{90.33}	& \textbf{94.80} & \textbf{79.00}	& \textbf{87.76}	\\
      & 50 & \textbf{99.43} & \textbf{94.56} & \textbf{94.27} & \textbf{93.72} & \textbf{89.58} & \textbf{94.31} & \textbf{79.14} & \textbf{86.97} \\
     & 25 & 98.22 & \textbf{88.40} & 89.07 & 90.75 & 81.87 & 89.66 & 	61.92 & 73.80 \\
    \end{tabular}
    \label{tab:original_prun}
    \vspace{-5mm}
\end{table*}

\subsection{Impact of Face Data Pruning}
To evaluate the effectiveness of our proposed DiffProb pruning strategy, we first conduct experiments under the same-model setups, i.e., when the dataset is pruned using a model with the same architecture and loss function as the one used for training. Specifically, a ResNet-50 model trained with CosFace loss on the full CASIA-WebFace dataset is used to compute prediction probabilities for our DiffProb and uncertainty scores for DynUnc. Models are then retrained from scratch on the pruned datasets with the same settings.

Table \ref{tab:original_prun} presents FR verification accuracies of FR models trained on the pruned dataset with our DiffProb as well as the ones achieved by applying DynUnc \cite{DBLP:conf/cvpr/HeY0Z22} and Rand using different pruning percentages of 25\%, 50\% and 75\%. For our DiffProb method, these pruning percentages are achieved by setting $t$ to 0.00003, 0.0008 and 0.01, respectively.
The first row in the table presents the results achieved by the model trained on the full dataset, i.e., 100\% of the training data.
Table~\ref{tab:original_prun} shows that DiffProb consistently outperforms baseline methods (DynUnc and Rand) when 25\% and 50\% of the data are pruned. Notably, with 25\% pruning, DiffProb even surpasses the full-data baseline by 0.15 percentage points on average across standard benchmarks and by a larger margin on the large-scale IJB-C benchmark (improving TAR@FAR of $10^{-5}$ and $10^{-4}$ by 16.05 and 2.5 percentage points, respectively). This demonstrates that DiffProb not only reduces dataset size but can also enhance performance through the removal of redundant or possibly noisy samples.



Figure~\ref{fig:pruning} illustrates samples from two different identities in the CASIA-WebFace dataset, displayed alongside their predicted probabilities $p(x_i)$. The samples are sorted from left to right in ascending order of $p(x_i)$. Translucent images indicate samples that were pruned by our proposed DiffProb method, while opaque images were retained. The figure shows that samples with highly similar $p(x_i)$ values are effectively identified and removed, reducing redundancy within each identity. Importantly, DiffProb preserves hard samples, those with low predicted probabilities, provided they are not redundant. This behavior, visible in the leftmost part of the figure, demonstrates DiffProb’s ability to retain informative and diverse training examples that could contribute meaningfully to the learning process.

\subsection{Generalizability Across Different Losses}
\label{sec:cross_loss}

\begin{table*}[]
    \centering
        \caption{Performance of FR models trained on datasets pruned with different methods to three pruning percentages. All the considered models use the same architecture (ResNet-50) as the FR model used to prune the dataset, but distinct loss functions, allowing for a detailed analysis of each method's effectiveness in the cross-loss scenario.
    For each training setting, the best result for each benchmark and pruning percentage is marked in bold. Note that our DiffProb achieved superior performances in the cross-loss setting.}
    \begin{tabular}{ccc||cccccc|cc}
    \multirow{2}{*}{\textbf{Loss Function}} & \multirow{2}{*}{\textbf{Method}} & \multirow{2}{*}{ \textbf{Kept Samples (\%)} } & \multirow{2}{*}{\textbf{LFW}} & \multirow{2}{*}{\textbf{CFP-FP}} & \multirow{2}{*}{\textbf{AgeDB}} & \multirow{2}{*}{\textbf{CA-LFW}} & \multirow{2}{*}{\textbf{CP-LFW}} & \multirow{2}{*}{\textbf{Avg}} & \multicolumn{2}{c}{\textbf{IJB-C}} \\
    & & & & & & & & & \textbf{$10^{-5}$} & \textbf{$10^{-4}$} \\ \hline 
     \multirow{10}{*}{CosFace} & - & 100 & 99.35	& 95.24 & 94.65 & 93.83 & 90.17 & 94.65 & 62.95	& 85.26 \\ \cline{2-11}
      & \multirow{3}{*}{DynUnc \cite{DBLP:conf/cvpr/HeY0Z22}} & 75 & 99.30 & 94.90 & 94.27 & 93.75 & 89.02 & 94.25 & 72.66 & 86.01 \\
     & & 50 & 99.18 & 92.43 & 93.35 & 93.20 & 86.77 & 92.99 & 67.90 & 82.56 \\
      & & 25 & \textbf{98.77} & 86.97 & \textbf{90.15} & \textbf{91.43} & \textbf{82.57} & \textbf{89.98} &	\textbf{66.56} & \textbf{78.16} \\ 
      \cline{2-11}
    & \multirow{3}{*}{Rand} & 75 & 99.40 & 94.54 & 94.22 & 93.27 & 89.33	& 94.15 & 75.74 & 86.22\\
     & & 50 & 99.30 & 93.30 & 93.40 & 92.83 & 87.97 & 93.36 & 75.69 & 84.59\\
      & & 25 & 98.38 & 87.50 & 89.22 & 90.87 & 81.53 & 89.50 & 62.13 & 74.77\\ 
     \cline{2-11}
     & \multirow{3}{*}{DiffProb (Ours)} & 75 & \textbf{99.50}	& \textbf{95.47}	& \textbf{94.82}	& \textbf{93.87}	& \textbf{90.33}	& \textbf{94.80} & \textbf{79.00}	& \textbf{87.76}	\\
     & & 50 & \textbf{99.43} & \textbf{94.56} & \textbf{94.27} & \textbf{93.72} & \textbf{89.58} & \textbf{94.31} & \textbf{79.14} & \textbf{86.97} \\
     & & 25 & 98.22 & \textbf{88.40} & 89.07 & 90.75 & 81.87 & 89.66 & 	61.92 & 73.80 \\
     \hline \hline
     \multirow{10}{*}{AdaFace} & - & 100 & 99.42 & 95.19 & 94.82 & 93.75	& 90.17	& 94.67 & 39.56 & 81.16\\ \cline{2-11}
     & \multirow{3}{*}{DynUnc \cite{DBLP:conf/cvpr/HeY0Z22}} & 75 & 99.37 & 94.31 & 94.25 & 93.40 & 89.15 & 94.10 & 74.51 & 86.65 \\
     & & 50 & 99.05 & 92.31 & 93.03 & 92.97 & 86.63 & 92.80 & 69.61 & 82.81 \\
      & & 25 & \textbf{98.55} & 87.27 & \textbf{90.48} & \textbf{91.32} & \textbf{82.43} & \textbf{90.01} & \textbf{64.10} & \textbf{77.38} \\
      \cline{2-11}
    & \multirow{3}{*}{Rand} & 75 & 99.30 & 94.77 & 94.15 & 93.40 & 89.22 & 94.17 & 72.63	& 85.04 \\
     & & 50 & 99.18 & 93.51 & 92.88 & 92.93 & 87.97 & 93.30 & 75.20 & 84.20 \\
      & & 25 & 98.53 & 86.99 & 88.88 & 90.42 & 81.63 & 89.29 & 61.27 & 73.80 \\ 
     \cline{2-11}
     & \multirow{3}{*}{DiffProb (Ours)} & 75 & \textbf{99.48}	& \textbf{95.41}	& \textbf{94.70} & \textbf{93.83} & \textbf{90.40}	& \textbf{94.77} & \textbf{77.86}	& \textbf{87.61}	\\
     & & 50 & \textbf{99.30} & \textbf{94.93} & \textbf{94.33} & \textbf{93.60} & \textbf{89.68} & \textbf{94.37} & \textbf{79.04} & \textbf{87.34}\\
     & & 25 & 98.02 & \textbf{87.73} & 88.65 & 90.55 & 81.93 & 89.38 &  61.35 & 72.76 \\
     \hline \hline
     \multirow{10}{*}{CurricularFace} & - & 100 & 99.48 &	95.54	& 95.00 &	93.82 &	89.68 &	94.71	&		1.97	& 22.59 \\ \cline{2-11}
     & \multirow{3}{*}{DynUnc \cite{DBLP:conf/cvpr/HeY0Z22}} & 75 & 99.42 & 94.61 & 94.45 & \textbf{93.60} & 89.15 & 94.25 & \textbf{71.52} & \textbf{86.34}\\
     & & 50 & 99.22 & 91.91 & 93.07 & 93.05 & 86.43 & 92.74 & 71.45 & 83.41 \\
      & & 25 & 98.58 & 87.26 & \textbf{89.88} & \textbf{91.30} & 82.55 & 89.91 & \textbf{67.56} & \textbf{78.30} \\
      \cline{2-11}
    & \multirow{3}{*}{Rand} & 75 & 99.32 & 94.39 & 94.25 & 93.47 & 89.17 & 94.12 & 7.97 & 45.96\\
     & & 50 & 99.33 & 93.51 & 93.03 & 92.87 & 87.30 & 93.21 & 41.93 & 79.42 \\
      & & 25 & 98.53 & 87.86 & 88.97 & 90.88 & 82.20 & 89.69 & 63.85 & 75.30\\ 
     \cline{2-11}
     & \multirow{3}{*}{DiffProb (Ours)} & 75 & \textbf{99.50}	& \textbf{95.06}	& \textbf{94.73}	& 93.58	& \textbf{89.53}	& \textbf{94.48} & 27.36	& 73.55\\
     & & 50 & \textbf{99.35} & \textbf{94.57} & \textbf{94.40} & \textbf{93.63} &  \textbf{89.07} & \textbf{94.20} & \textbf{73.51} & \textbf{86.11} 	\\
     & & 25 & \textbf{98.70} & \textbf{89.54} & 89.57 & 91.07 & \textbf{83.58} & \textbf{90.49} & 62.80 & 75.49\\
    \end{tabular}
    \label{tab:local_prun_cross_loss}
\end{table*}

To assess generalization across different loss functions, we evaluate DiffProb in a cross-loss setting where pruned datasets are used to train models with different loss functions (AdaFace and CurricularFace) while keeping the ResNet-50 architecture fixed. These experiments verify whether pruned data produced by DiffProb remains effective beyond the conditions under which it was generated.
Specifically, we trained 18 instances of ResNet-50 from scratch with our pruned datasets by our DiffProb (6 instances), DynUnc (6 instances), and Rand (6 instances) with pruning ratios of 25\%, 50\% and 75\%. 
For each pruning ratio and method, the pruning process is performed once and the resultant pruned dataset is used to train models with AdaFace and CurricularFace.
It can be clearly observed that DiffProb consistently outperforms DynUnc \cite{DBLP:conf/cvpr/HeY0Z22} and random pruning when pruning 25\% and 50\% of the data, following similar observations to the ones discussed in the previous section. This difference in performance is particularly noticeable when 50\% of the dataset is pruned since DiffProb largely maintains the performance levels of the original network while the remaining methods fall significantly behind. As an example, using 50\% of the dataset to train a ResNet-50 \cite{He2015DeepRL} with the AdaFace loss results in a significant decrease of 1.87 and 1.37 percentage points for DynUnc \cite{DBLP:conf/cvpr/HeY0Z22} and random, respectively, when compared with the original model, while leading to a marginal decrease of 0.30 percentage points when using our DiffProb. A similar observation is also verified for the large-scale benchmark IJB-C, where the proposed DiffProb is even able to surpass models trained on the complete dataset. In particular, our method consistently surpasses the original model when 50\% of the data is pruned while resulting in a reduction of training time by a factor of two. Similarly to what was concluded in the previous section, pruning 75\% of the data leads to significant performance drops in all the considered settings, highlighting the challenge of training FR with such a small dataset.
These results validate the robustness of our DiffProb pruning strategy across diverse training objectives.


\subsection{Generalizability Across Different Network Architectures}
\label{sec:cross_net}

We further explore the effectiveness of DiffProb in a cross-architecture setting by training models with ResNet-34 using the datasets pruned based on the predictions of a ResNet-50 pre-trained with the CosFace loss. Specifically, we fix the loss function as the same considered by the pre-trained FR model used to perform the pruning (CosFace \cite{wang2018cosfacelargemargincosine}) while varying the network architecture (ResNet-34 \cite{He2015DeepRL}). 

The achieved verification accuracies are reported in Table \ref{tab:local_prun_cross_net}. 
One can observe that DiffProb again leads to minimal verification accuracy loss at 25\% and 50\% pruning ratios and outperforms competitors (Rand and DynUnc) across most benchmarks.
In particular, our DiffProb leads to a marginal decrease in average performance on the small benchmarks when removing half of the data (0.22 percentage points) while significantly surpassing the model trained on the full dataset on IJB-C, supporting the conclusions withdrawn in the previous section. These results suggest that our DiffProb allows to prune half of the data and consequently reduce the training time by half without significantly compromising the FR performances. In contrast, both DynUnc and random pruning degrade performance more substantially. These findings demonstrate the architectural generalizability of DiffProb and highlight its practical value in real-world deployment scenarios where retraining with different model architectures is common.


\begin{table*}[ht!]
    \centering
        \caption{Accuracies of FR models trained on datasets pruned with different methods to three pruning percentages. All the considered modls use the same loss function (CosFace) as the FR model used to prune the original dataset, but distinct architectures, allowing for a detailed analysis of each method's effectiveness in the cross-network scenario.
    For each training setting, the best result for each benchmark and pruning percentage is marked in bold. Note that our DiffProb provides leading performance in the cross-network scenario.}
    \begin{tabular}{ccc||cccccc|cc}
    \multirow{2}{*}{\textbf{Network}} & \multirow{2}{*}{\textbf{Method}} & \multirow{2}{*}{ \textbf{Kept Samples (\%)} } & \multirow{2}{*}{\textbf{LFW}} & \multirow{2}{*}{\textbf{CFP-FP}} & \multirow{2}{*}{\textbf{AgeDB}} & \multirow{2}{*}{\textbf{CA-LFW}} & \multirow{2}{*}{\textbf{CP-LFW}} & \multirow{2}{*}{\textbf{Avg}} & \multicolumn{2}{c}{\textbf{IJB-C}} \\
    & & & & & & & & & \textbf{$10^{-5}$} & \textbf{$10^{-4}$} \\ \hline 
     \multirow{10}{*}{ResNet-50} & - & 100 & 99.35	& 95.24 & 94.65 & 93.83 & 90.17 & 94.65 & 62.95	& 85.26 \\ \cline{2-11}
     & \multirow{3}{*}{DynUnc \cite{DBLP:conf/cvpr/HeY0Z22}} & 75 & 99.30 & 94.90 & 94.27 & 93.75 & 89.02 & 94.25 & 72.66 & 86.01 \\
     & & 50 & 99.18 & 92.43 & 93.35 & 93.20 & 86.77 & 92.99 & 67.90 & 82.56 \\
      & & 25 & \textbf{98.77} & 86.97 & \textbf{90.15} & \textbf{91.43} & \textbf{82.57} & \textbf{89.98} &	\textbf{66.56} & \textbf{78.16} \\ 
      \cline{2-11}
    & \multirow{3}{*}{Rand} & 75 & 99.40 & 94.54 & 94.22 & 93.27 & 89.33	& 94.15 & 75.74 & 86.22\\
     & & 50 & 99.30 & 93.30 & 93.40 & 92.83 & 87.97 & 93.36 & 75.69 & 84.59\\
      & & 25 & 98.38 & 87.50 & 89.22 & 90.87 & 81.53 & 89.50 & 62.13 & 74.77\\ 
     \cline{2-11}
     & \multirow{3}{*}{DiffProb (Ours)} & 75 & \textbf{99.50}	& \textbf{95.47}	& \textbf{94.82}	& \textbf{93.87}	& \textbf{90.33}	& \textbf{94.80} & \textbf{79.00}	& \textbf{87.76}	\\
     & & 50 & \textbf{99.43} & \textbf{94.56} & \textbf{94.27} & \textbf{93.72} & \textbf{89.58} & \textbf{94.31} & \textbf{79.14} & \textbf{86.97} \\
     & & 25 & 98.22 & \textbf{88.40} & 89.07 & 90.75 & 81.87 & 89.66 & 	61.92 & 73.80 \\
     \hline \hline
     \multirow{10}{*}{ResNet-34} & - & 100 & 99.37 &	94.97&	94.35&	93.47&	89.45&	94.32&62.04&	84.84\\ \cline{2-11} 
     & \multirow{3}{*}{DynUnc \cite{DBLP:conf/cvpr/HeY0Z22}} & 75 & 99.27 & 94.07 & 93.98 & 93.50 & 88.50 & 93.86 & 73.45 & 85.45\\
     & & 50 & 99.15 & 91.61 & 92.80 & 92.93 & 86.52 & 92.60 & 71.85 & 83.43\\
      & & 25 & \textbf{98.68} & 86.33 & \textbf{89.60} & \textbf{91.70} & \textbf{82.33} & \textbf{89.73} & \textbf{64.57} & \textbf{76.92} \\ 
      \cline{2-11}
    & \multirow{3}{*}{Rand} & 75 & 99.32 & 94.44 & 93.88 & 93.40 & 88.60 & 93.93 & 72.93	& 85.19 \\
     & & 50 & 99.23 & 93.27 & 93.28 & 92.62 & 87.10 & 93.10 & 74.44 & 83.28 \\
      & & 25 & 98.17 & 86.84 & 88.58 & 90.45 & 81.52 & 89.11 & 63.09 & 74.02 \\ 
     \cline{2-11}
     & \multirow{3}{*}{DiffProb (Ours)} & 75 & \textbf{99.38}	& \textbf{95.06}	& \textbf{94.47} & \textbf{93.68}	& \textbf{89.77}	& \textbf{94.47} & \textbf{78.40} & \textbf{87.22} \\
     & & 50 & \textbf{99.32} & \textbf{94.67} & \textbf{94.25} & \textbf{93.50} & \textbf{88.78} & \textbf{94.10} & \textbf{78.66} & \textbf{86.56} \\
      & & 25 & 98.30 & \textbf{87.57} & 88.62 & 90.83 & 81.77 & 89.42 & 60.01 & 73.09\\ 
    \end{tabular}
    \label{tab:local_prun_cross_net}
\end{table*}

\begin{table*}[ht!]
    \centering
        \caption{Performance of FR models trained on datasets generated with and without guidance of our auxiliary cleaning mechanism. The first two rows present the results of training the model on the complete dataset or its cleaned version, without applying any pruning. The second part of the table refers to our DiffProb method applied alone or after the pre-processing step introduced by our cleaning mechanism. The best result for each benchmark and pruning percentage is marked in bold. Note that both the cleaning and the pruning operations provide additional performance advantages in most experimental setups.}
    \begin{tabular}{cc||cccccc|cc}
    \multirow{2}{*}{\textbf{Method}} & \multirow{2}{*}{ \textbf{Kept Samples (\%)} } & \multirow{2}{*}{\textbf{LFW}} & \multirow{2}{*}{\textbf{CFP-FP}} & \multirow{2}{*}{\textbf{AgeDB}} & \multirow{2}{*}{\textbf{CA-LFW}} & \multirow{2}{*}{\textbf{CP-LFW}} & \multirow{2}{*}{\textbf{Avg}} & \multicolumn{2}{c}{\textbf{IJB-C}} \\
    & & & & & & & & \textbf{$10^{-5}$} & \textbf{$10^{-4}$} \\ \hline 
     - & 100 & 99.35	& \textbf{95.24} & 94.65 & \textbf{93.83} & 90.17 & 94.65 & 62.95	& 85.26 \\ \hline 
      Clean & 98.87 & \textbf{99.47} &	95.21& \textbf{94.85}& 	93.68	& \textbf{90.33}& 	\textbf{94.71}	& \textbf{75.02} & \textbf{86.82}\\ \hline \hline
    \multirow{3}{*}{DiffProb (Ours)} & 75 & \textbf{99.50}	& \textbf{95.47}	& 94.82	& \textbf{93.87}	& 90.33	& 94.80 & 79.00	& 87.76	\\
     & 50 & \textbf{99.43} & 94.56 & 94.27 & \textbf{93.72} & 89.58 & 94.31 & 79.14 & 86.97 \\
      & 25 & 98.22 & 88.40 & \textbf{89.07} & 90.75 & \textbf{81.87} & 89.66 & 	\textbf{61.92} & 73.80 \\ \hline
      \multirow{3}{*}{Clean + DiffProb (Ours)} &	75 &  99.40	& 95.40	& \textbf{94.83}	& 93.85 &	\textbf{90.55}	& \textbf{94.81} &  \textbf{79.34}  & \textbf{87.79}\\
      & 50 & 99.33	& \textbf{94.93}	& \textbf{94.53}	& 93.62	& \textbf{89.80}	& \textbf{94.44} & \textbf{79.27} & \textbf{87.39} \\
      & 25 & \textbf{98.33}	& \textbf{88.80}	& 88.93	& \textbf{91.13}	& 81.55	& \textbf{89.75} & 61.89 & \textbf{74.01} \\
    \end{tabular}
    \label{tab:local_prun_clean}
    \vspace{-5mm}
\end{table*}

\subsection{Impact of Data Cleaning}
\label{sec:res_cleaning}
\begin{figure}[thpb]
    \centering
    \includegraphics[width=0.7\linewidth]{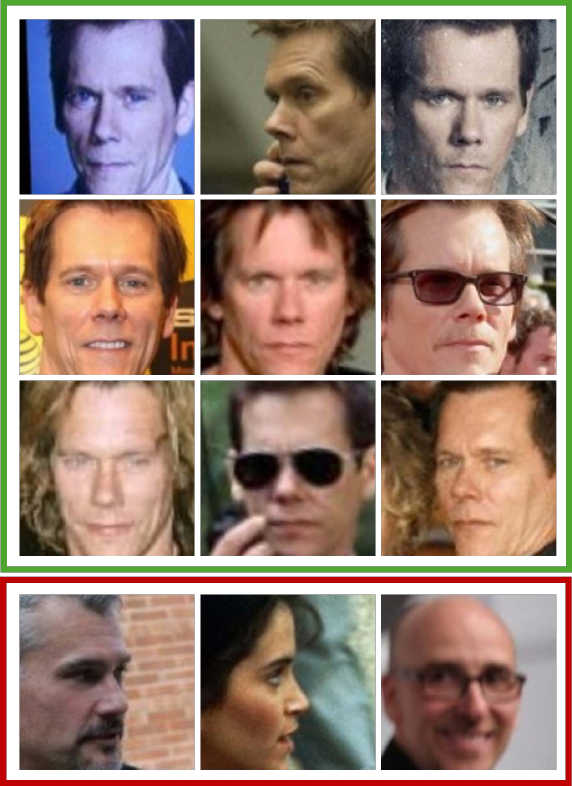}
    \vspace{-3mm}
    \caption{A set of samples categorized as belonging to the same identity on the CASIA-WebFace dataset \cite{DBLP:journals/corr/YiLLL14a}. The red box highlights the three samples considered as mislabeled or label-flipped by our auxiliary data cleaning mechanism. The green box contains examples of samples that were not removed by our data cleaning approach.}
    \label{fig:cleaning}
    \vspace{-2mm}
\end{figure}

To isolate the impact of noisy labels, we evaluate the contribution of our auxiliary cleaning mechanism, which removes mislabeled or label-flipped samples prior to pruning.
Figure \ref{fig:cleaning} shows samples labeled as belonging to the same identity according to the CASIA-WebFace dataset ground truth labels \cite{DBLP:journals/corr/YiLLL14a}. The images inside the green box constitute examples of samples considered as correctly labeled by our auxiliary data cleaning method, while the red box encompasses the three samples removed through data cleaning. It can be visually observed that none of the latter samples belong to the claimed identity, showing that our auxiliary data cleaning method could effectively remove mislabeled and label-flipped samples.

Table \ref{tab:local_prun_clean} presents the results of our proposed DiffProb pruning mechanism applied with and without the extra guidance of our auxiliary cleaning mechanism. The proposed cleaning process removes 1.13\% of the samples on CASIA-WebFace \cite{DBLP:journals/corr/YiLLL14a}. The comparison between the first two lines of Table \ref{tab:local_prun_clean} shows that removing such a small number of entries increases the performance of the FR model. A particularly high performance difference is verified when evaluating on the large benchmark, IJB-C, at a threshold of $10^{-5}$, where cleaning the data increased the TAR@FAR by 14.07 percentage points. Similar conclusions can be drawn when comparing the results obtained by using our DiffProb method alone or by applying it on top of the already clean dataset. Data cleaning before pruning generally leads to higher verification performance, suggesting that the proposed auxiliary cleaning mechanism successfully removes harmful samples that are kept when directly pruning the data. 
The results achieved by the model trained on clean and pruned data (noted as Clean+DiffProb) surpass the model trained on only pruned data (DiffProb). This confirms the complementary benefit of combining cleaning and pruning to maximizing training efficiency and accuracy.

\section{Conclusion}
\label{sec:conclusion}

This work is the first to propose
and evaluate a data pruning approach,
DiffProb, for the FR task.
By analyzing prediction probabilities of samples within each identity, DiffProb identifies and removes samples that are likely to contribute similarly to FR model training, allowing for substantial dataset reduction with up to 50\% and without sacrificing FR performance and even, in some settings, surpassing an FR model trained on the complete dataset. 
We further evaluate the generalizability of DiffProb
under cross-loss and cross-network training settings, demonstrating the robustness of our approach.
We also compared DiffProb with previous data pruning methods designed for common computer vision tasks,  
achieving superior results in most of the considered settings. 
To eliminate the impact of noisy labels, we also proposed an auxiliary cleaning mechanism that effectively detects and removes mislabeled or label-flipped samples. When combining data cleaning with DiffProb, the FR verification accuracies are further improved, especially on large-scale benchmarks.
Finally, this paper presents the first principled and generalizable approach to reduce the burden of massive FR datasets, enabling FR training with few annotated face data, lowering computational costs, and minimizing the risks associated with storing and processing large volumes of sensitive biometric data. Future research could explore identity-level pruning to align more closely with privacy-aware training paradigms.






\section*{Ethical Impact Statement and Limitations}
This is the first work proposing face data pruning to enable training face recognition models with few annotated face data. This is mainly driven by technical limitations of collecting, sharing and storing sensitive biometric data as well as the respective legal and ethical concerns \cite{GDPR_practice, bipa}. We recognize that future research work could consider pruning complete identities, reducing the number of user consents \cite{GDPR_practice, bipa} needed to collect, store and share sensitive biometric data \cite{lirias3838501}. At the same time, we acknowledge but firmly reject the potential for malicious or unlawful use of this and similar machine learning technologies. Misuse of face recognition may involve processing biometric data outside legal boundaries and without individuals’ consent, for purposes such as user profiling or unauthorized functionality beyond identity verification \cite{DBLP:journals/tifs/MedenRTDKSRPS21}.

{\small
\bibliographystyle{ieee}
\bibliography{main}
}

\end{document}